\documentclass[10pt]{article}
\usepackage[preprint]{tmlr}
\usepackage{amsmath,amssymb}
\usepackage{graphicx}
\usepackage{booktabs,tabularx,array,xltabular}
\usepackage{enumitem}
\setlist{itemsep=1pt,topsep=3pt,parsep=0pt}

\usepackage{hyperref}
\hypersetup{colorlinks=true,linkcolor=black,citecolor=blue!50!black,urlcolor=blue!50!black}
\usepackage{xcolor}
\usepackage{url}
\usepackage{xurl}

\title{Do Time-Series Foundation Models Forecast Index-Futures Returns? A Preregistered Evaluation of Models, Metrics, and Input Representations}
\author{\name Rishabh Upadhyay \email uhrishabh@gmail.com \\
        \addr Inception 42, Abu Dhabi, United Arab Emirates}

\begin{document}
\maketitle
\begin{abstract}
We ask whether the apparent forecasting skill of frozen time-series foundation models (TSFMs) on intraday index-futures returns is incremental skill. In a pre-registered evaluation on E-mini Nasdaq-100 (NQ) and E-mini S\&P 500 (ES) futures, a common panel of five zero-shot models (Chronos-Bolt, Chronos-2, TimesFM 2.5, Moirai 2.0 and TiRex), plus separate experiments with Kronos, is compared with a no-change forecast at 5-120 minutes, using native prices, log returns, volatility-standardised returns and a volume clock as inputs. On a blind set opened once, 0 of 160 model-representation-horizon-market cells beat no-change, and a Hansen SPA test gives p = 1.000 on both markets; a later retrospective hold-out, scored as a descriptive check, gives 0 of 120. Input representations change forecast magnitude but not accuracy, and a transformation-aware ridge regression does as well. Several standard measures nevertheless suggest skill. Price-level correlation exceeds 0.9996, but no-change scores higher. Within-day RankIC across forecast origins rewards a trivial revert-to-mean rule (+0.35 at 30 minutes); we derive the expected within-day covariance under a random walk in closed form, and a sign-randomised null reproduces about 95\% of the observed value. One model beats HAR for 5-minute volatility, but not the EWMA its input already contains. On strongly planted synthetic signals the models recover part of the signal at 5 minutes, and recovery deteriorates at 30 minutes. Benchmark-level success of TSFMs should therefore not be assumed to carry over to low-signal domains such as intraday returns. An automated reproducibility audit rebuilds every reported number from the saved forecasts.

\end{abstract}

\section{Introduction}\label{sec:1}
Foundation models for time series promise zero-shot forecasting across domains, and finance-specific ones such as Kronos report large gains on rank-correlation metrics \citep{shi2026kronos}. Whether such models forecast short-horizon returns of liquid futures better than the random walk is an empirical question with a strong prior, because intraday index-futures returns are close to unpredictable. A credible test therefore needs a benchmark that is hard to beat (the no-change forecast), data the models are unlikely to have seen, a fixed analysis plan so that many models, horizons and input variants do not turn noise into a finding, and controls that separate model skill from simple ingredients.

We pre-registered such a test on 1-minute data for E-mini Nasdaq-100 (NQ) and E-mini S\&P 500 (ES) futures. It is organised around one question: is apparent skill incremental skill? The same forecasts look skilful under several standard measures, namely level correlation, within-day RankIC and beating HAR for volatility. Under matched nulls, ingredient controls and a magnitude-sensitive loss, none of them shows incremental skill on real returns. The contributions are:

\begin{enumerate}
\item \textbf{A pre-registered blind evaluation} of six TSFM families, a common frozen panel of five (Chronos-Bolt, Chronos-2, TimesFM 2.5, Moirai 2.0, TiRex) plus Kronos, with a confirmatory multiplicity test (Hansen SPA) and a pre-set smallest meaningful improvement. On the blind set 0/160 cells beat no-change at 5-120 minutes, and a later retrospective hold-out, scored descriptively, gives 0/120.
\item \textbf{Input-representation experiments under a fixed protocol} (log returns, volatility-standardised returns and a volume clock), with a transformation-aware ridge comparator. Representations shrink or inflate forecasts, but none creates skill.
\item \textbf{Controls that explain the apparent successes.} Price-level correlation exceeds 0.9996 for every model and is higher for no-change. The volatility gain over HAR is reproduced by the causal EWMA that the representation injects. Within-day RankIC across forecast origins is positive under a random walk for a revert-to-mean predictor; we derive the expected covariance in closed form and calibrate the rank metric with a sign-randomised null.
\item \textbf{Explicit detection limits.} Planted-signal power curves and a points bound for a fixed trading policy state which effects the study can and cannot exclude. On a strong planted signal the models recover part of it at 5 minutes, and recovery deteriorates at 30 minutes (\S\ref{sec:5.6}).
\end{enumerate}
We do not claim that returns are unpredictable in general, that these models cannot learn financial structure, or that any published number is invalid. Section~\ref{sec:7} discusses what the results imply for adopting foundation models in low-signal domains.

\section{Related work}\label{sec:2}
\textbf{Time-series foundation models.} Pretrained zero-shot forecasters include Chronos and Chronos-Bolt \citep{ansari2024chronos}, Chronos-2 \citep{ansari2025chronos2}, TimesFM \citep{das2024timesfm}, Moirai \citep{woo2024moirai}, Moirai 2.0 \citep{liu2025moirai2} and TiRex \citep{auer2025tirex}. Kronos is pretrained on more than 12 billion financial K-line records \citep{shi2026kronos}.

\textbf{Financial evaluations of TSFMs.} Recent studies reach cautious conclusions. \citet{noguer2026} compare TimesFM 2.5, Moirai 2.0, Chronos and Chronos-2 with random-walk and trained baselines on individual equity returns and report gains that are small and sparse. \citet{zhong2026} find that foundation models do not beat simple volatility benchmarks on 5,000 Chinese stocks. \citet{brini2026} finds that recalibration removes much of the apparent advantage of nine TSFMs over HAR-family models for realised volatility; our EWMA control plays the same role for intraday futures. \citet{konat2026} find that four pretrained TSFMs do not outperform the random walk on 19 fixed-income and FX instruments. \citet{rahimikia2025} find off-the-shelf TSFMs weak for daily cross-sectional excess returns, and \citet{marconi2025} finds zero-shot TTM and Chronos competitive on some financial tasks but matched by specialised models. \citet{wan2026collapse} document ``forecast collapse'', flattened and poorly ranked forecasts of hourly equity returns, and trace it to low predictability and per-series objectives. Our models show both regimes: Kronos-small's forecasts are about 1.5x as dispersed as the realised moves, while the representations that shrink forecasts toward zero are the ones that reduce error. CombinationTS \citep{wang2026combinationts} attributes forecasting performance to modular components, including input transformations; our representation study asks a narrower, pre-registered version of that question for returns. The table below positions this study against the closest work.

{\footnotesize\par\medskip\noindent
\begin{tabularx}{\linewidth}{@{}>{\raggedright\arraybackslash}X>{\raggedright\arraybackslash}X>{\raggedright\arraybackslash}X>{\raggedright\arraybackslash}X>{\raggedright\arraybackslash}X@{}}
\toprule
Study & Target, frequency & Benchmark & Input / model variants & Holdout design \\
\midrule
\citet{noguer2026} & five US equities, linear and log returns & random walk, trained nets & native & rolling-origin \\
\citet{wan2026collapse} & 1,000 equities, hourly returns, cross-section & ranking and calibration & training objectives & - \\
\citet{brini2026} & realised volatility, 50 assets & HAR family & recalibration & - \\
\citet{konat2026} & 19 fixed-income and FX instruments & random walk & four pretrained TSFMs & - \\
\citet{zhong2026} & volatility, 5,000 Chinese stocks & simple volatility benchmarks & incl. fine-tuning & - \\
\textbf{This study} & \textbf{index futures, 5-120 min and daily returns} & \textbf{no-change, ridge, HAR, EWMA, random-direction nulls} & \textbf{R0/R1/R2/R4, frozen; Kronos also trained and fine-tuned} & \textbf{pre-registered blind + retrospective hold-out} \\
\bottomrule
\end{tabularx}
\par\medskip}

We do not claim to be the first to find TSFM weakness on financial returns, input-transformation effects or volatility-calibration effects. The contribution is a controlled, pre-registered combination of them on a hard, liquid target, with explicit detection limits.

\textbf{Metrics and mechanical predictability.} Kronos reports IC and RankIC; for price-series forecasting these are computed per sample across the forecast horizon's steps, and for return forecasting they are pooled over samples \citep[App.~D]{shi2026kronos}. FinTSB \citep{fintsb2025} defines IC per asset over time and then averages over assets. Our within-day metric differs from both (Section~\ref{sec:6}). Evaluation statistics can be predictable or inflated by construction: \citet{mccarthy2025} show this for the directional accuracy of temporal aggregates, \citet{ahn2026} for overlapping returns in time-series momentum, and earlier work for random-walk detrending \citep{nelson1981} and small-sample autocorrelation \citep{kendall1954}. The bias behind Section~\ref{sec:6} is classical: \citet{kendall1954} on the sample autocorrelation, \citet{dickey1979} on the regression with a constant under a unit root, and \citet{nickell1981} for a within-group analogue. Our within-day RankIC result is a specific instance of this family. To our knowledge, this construction, ranking across forecast origins that share one day's path, has not been documented before, although the underlying phenomenon is not new.

\section{Data, splits and pre-registration}\label{sec:3}
\textbf{Markets, origins and target.} We use NQ and ES front-month futures: 1-minute OHLCV bars for all hours, built from Databento CME Globex data (1-second bars for the volume clock) and rolled 8 days before the third Friday. NQ is the primary market and ES the replication market. Intraday origins t are fixed clock times in US Eastern time: every 30 minutes from 10:00 to 15:30 for the 30-minute horizon (12 per day), 10:00-13:30 for 120 minutes (8 per day), and analogous grids at 5 and 60 minutes. Windows that cross a roll or a session gap longer than 60 minutes are dropped. The target is $y=\log\bigl(P(t+h)/P(t)\bigr)$, where P(t) is the opening (first) trade of minute t and h $\in$ \{5, 30, 60, 120\} minutes. Every forecaster, every representation, no-change and ridge are scored against this one target column.

\textbf{Information set.} Every foundation-model forecast f is a log return anchored at the last context close, close(t-1). Contexts contain completed bars only, ending with minute t-1, and f is compared directly with y without re-anchoring. The transformation-aware ridge comparator also uses bars up to t-1 only and therefore has exactly the models' information set. Only two validation-stage baselines (B2 ridge, B3 LightGBM) also use P(t); they appear in no blind comparison. close(t-1) and open(t) coincide for 26\% of NQ and 55\% of ES origins. In a post-hoc sensitivity analysis (not pre-registered; added after an external review) we rescored all 160 blind cells against a close-anchored target, $\log c(t{+}h{-}1)-\log c(t{-}1)$: again 0/160 beat no-change (best ratio 1.0075 vs 1.0073), and the median change in the ratio is 0.0000. Appendix~\ref{app:C} gives the per-model decoding.

\textbf{Daily sample.} The daily hypotheses (H9, H11) use a separate sample with one origin per session, built from a Panama roll-adjusted daily series, and next-session returns as targets. Blind has NQ 349 sessions (2025-01-02 to 2026-05-08) and ES 332; the daily hold-out has NQ 119 (2026-05-12 to 2026-10-29) and ES 136 (2026-04-17 to 2026-10-29). Appendix~\ref{app:D} explains why these counts differ from the intraday sample.

\textbf{Splits.} All splits were fixed before any model was run.

{\footnotesize\par\medskip\noindent
\begin{tabularx}{\linewidth}{@{}l>{\raggedright\arraybackslash}X>{\raggedright\arraybackslash}X>{\raggedright\arraybackslash}X@{}}
\toprule
Set & NQ & ES & Role \\
\midrule
Train & 2018-01 - 2022-12 & 2020-01 - 2022-12 & fitting baselines; Kronos trained from scratch \\
Validation & 2023-01 - 2024-12 & 2023-01 - 2024-12 & all development choices \\
Blind & 2025-01-01 - 2026-05-11 (333 days) & 2025-01-01 - 2026-04-16 (316 days) & opened once, after the freeze \\
Hold-out (``ultra-blind'') & 2026-09-03 - 2026-10-30 (39 days) & same & final descriptive check \\
\bottomrule
\end{tabularx}
\par\medskip}

The hold-out is TopstepX 1-minute history recorded in Sep-Oct 2026, before the plan was written (2026-11-02), and not opened by this study until 2026-11-04. It is therefore a retrospective hold-out, not prospectively collected data. The same NQ capture had been used by the authors in unrelated trading-strategy work, but never with these models, their forecasts, or any choice in this study.

\textbf{Contamination.} Model release dates and published cutoffs were recorded before any run (PLAN \S{}10). Blind is after Kronos's published cutoff (June 2024) and after Chronos-Bolt's release; for the models released during 2025 only part of Blind is guaranteed clean. The hold-out postdates every release, and a period-split check found no advantage on pre-cutoff years for any model (\S\ref{sec:5.5}).

\textbf{Pre-registration.} ``Pre-registered'' here means a locally time-logged and SHA-256-hashed analysis plan, not a third-party registration. The plan was frozen on 2026-11-02, and validation development followed with every amendment hashed. The confirmatory family, the daily hypotheses and the code hashes were locked on 2026-11-04 at 03:49 UTC; the blind set was opened once at 09:02 UTC and the hold-out at 17:33 UTC the same day. ``Blind'' means that no per-origin outcome or forecast on those sets was seen before scoring; aggregate return statistics (count, mean and standard deviation) were printed by the data builders before the freeze. Appendix~\ref{app:D} gives the full chronology and Section~\ref{sec:8} the resulting limitations.

\section{Models, representations and evaluation protocol}\label{sec:4}
\textbf{Models and baselines.} All foundation models are frozen and zero-shot (exact revisions in the release manifest): Kronos-small with Kronos-Tokenizer-base (24.7M parameters), Chronos-Bolt-base (205M), Chronos-2 (about 120M), TimesFM 2.5 (200M), Moirai 2.0-R-small (11.4M) and TiRex (about 35M). Each receives the last 400 completed 1-minute bars, and the point forecast is the median of 20 sampled paths (Kronos) or of the predicted quantiles. The baselines are no-change (f = 0), ridge regression on recent returns, LightGBM on bar features and, for volatility, HAR \citep{corsi2009} and a causal EWMA. Kronos trained from scratch on Train and Kronos-small fine-tuned on Train serve as training-based controls.

\textbf{Input representations.} The five-model panel receives four inputs; Kronos stays on native candlesticks because its tokenizer expects OHLCV. \textbf{R0} is the native price series. \textbf{R1} is 1-minute log returns, decoded as the sum of per-step median returns over h steps. \textbf{R2} is causally volatility-standardised returns, $z_k=r_k/\sigma_k$, where $\sigma_k$ is an EWMA ($\alpha$ = 0.05) of past squared returns; the inversion holds $\sigma_{t+1|t}$ fixed over the horizon. \textbf{R4} is a volume clock: bars close each time cumulative volume reaches the median 1-minute volume over the fit years (NQ 111, ES 321 contracts), the context is the last 400 volume bars completed by t, and the number of future steps $m=\max(1,\mathrm{round}(h\cdot\mathrm{rate}_t))$ is set from the past 60 minutes' bar rate, while the target stays the clock-time return. TimesFM is capped at 1,024 steps, so its R4 coverage is partial at 60-120 minutes. A multi-resolution input (R3) was secondary, and signed order flow (R5) was excluded because no signed trade data were available. The comparator for R1 and R2 is a transformation-aware ridge regression of y on the last 20 values of the same transformed input plus 5/15/60-step sums, fit on 2018-2021 with $\alpha$ chosen on 2022; R4 has no ridge comparator. Appendix~\ref{app:D} lists which model enters which analysis.

\textbf{Accuracy and the smallest meaningful improvement.} Return accuracy is measured by a loss on the predicted return, not by a direction-classification metric. The loss is the MAE ratio against no-change, $\sum|y-f|/\sum|y|$, with 95\% day-block bootstrap intervals; values below 1 favour the model. A cell \textbf{beats no-change} only if its ratio is $\le$ 0.995 and the upper bound of its interval is \textless{} 1 (the smallest meaningful improvement, SMI). A representation \textbf{improves on native} only if $\mathrm{ratio}_R-\mathrm{ratio}_{R0}$ $\le$ -0.005 with upper bound \textless{} 0, and a cell \textbf{adds over ridge} only if it is below ridge with a paired interval excluding 0.

\textbf{Confirmatory family.} The family C1 is \{R1, R2, R4\} $\times$ 5 core models $\times$ \{5, 30, 60, 120\} min = 60 cells per market. Hansen's SPA test \citep{hansen2005} of ``no cell beats no-change'' uses the consistent p-value, the stationary bootstrap \citep{politis1994} over trading days with mean block 1 day and 1,000 resamples, and the loss differential $d=|y|-|y-f|$. Missing origins (TimesFM R4) are handled available-case as registered, with a common-sample sensitivity analysis. The conclusion ``representation rescues frozen TSFMs'' required SPA to reject on NQ, at least one cell to meet the SMI against both no-change and R0 on NQ, and the same sign on ES; otherwise the conclusion is the bounded negative. Diebold-Mariano tests \citep{diebold1995} on squared error are used only descriptively and inside the power criterion, never as a test of MAE superiority; Appendix~\ref{app:C} lists the loss behind every test.

\textbf{Volatility (H7).} The registered rule requires QLIKE \citep{patton2011} below HAR, a QLIKE Diebold-Mariano p \textless{} 0.05, and 80\% band coverage within 75-85\%. A stricter volatility SMI (PLAN amendment 13.5) requires a QLIKE ratio $\le$ 0.98 with bootstrap upper bound \textless{} 1. The causal EWMA is reported alongside as an ingredient control.

\textbf{Daily hypotheses.} Two hypotheses were generated on Validation and registered before Blind. \textbf{H9:} Kronos-small's daily forecast, with its rolling 60-session bias removed, ranks next-session NQ returns (one-sided Spearman p \textless{} 0.05), and trading its sign (1 MNQ, \$2/pt, 0.85 pt round-trip cost) beats both zero and the 95th percentile of 200 random-direction draws. \textbf{H11:} TiRex's daily forecast, faded, does the same. Both are judged on NQ Blind. ES and the hold-out are reported whatever they show, with the registered rule and its 0.85 pt cost applied unchanged.

\textbf{Detection power (H12).} Synthetic series keep the real sequence of $|r_k|$ from Validation but replace the signs with a Markov chain that repeats the previous sign with probability 0.5 + $\delta$. Detection requires all three registered conditions: MAE ratio \textless{} 1, a one-sided squared-error Diebold-Mariano p \textless{} 0.05, and Spearman IC \textgreater{} 0 with p \textless{} 0.05. The pre-registered grid was $\delta$ $\le$ 0.05. It was extended to \{0.1, 0.2, 0.3\} after the oracle itself failed to detect $\delta$ $\le$ 0.05 (disclosed, PLAN \S{}13.10) and replicated 11 times (PLAN \S{}13.11).

\section{Results}\label{sec:5}
\subsection{Return forecasts: no detected skill}\label{sec:5.1}
\begin{figure}[t]\centering\includegraphics[width=\linewidth]{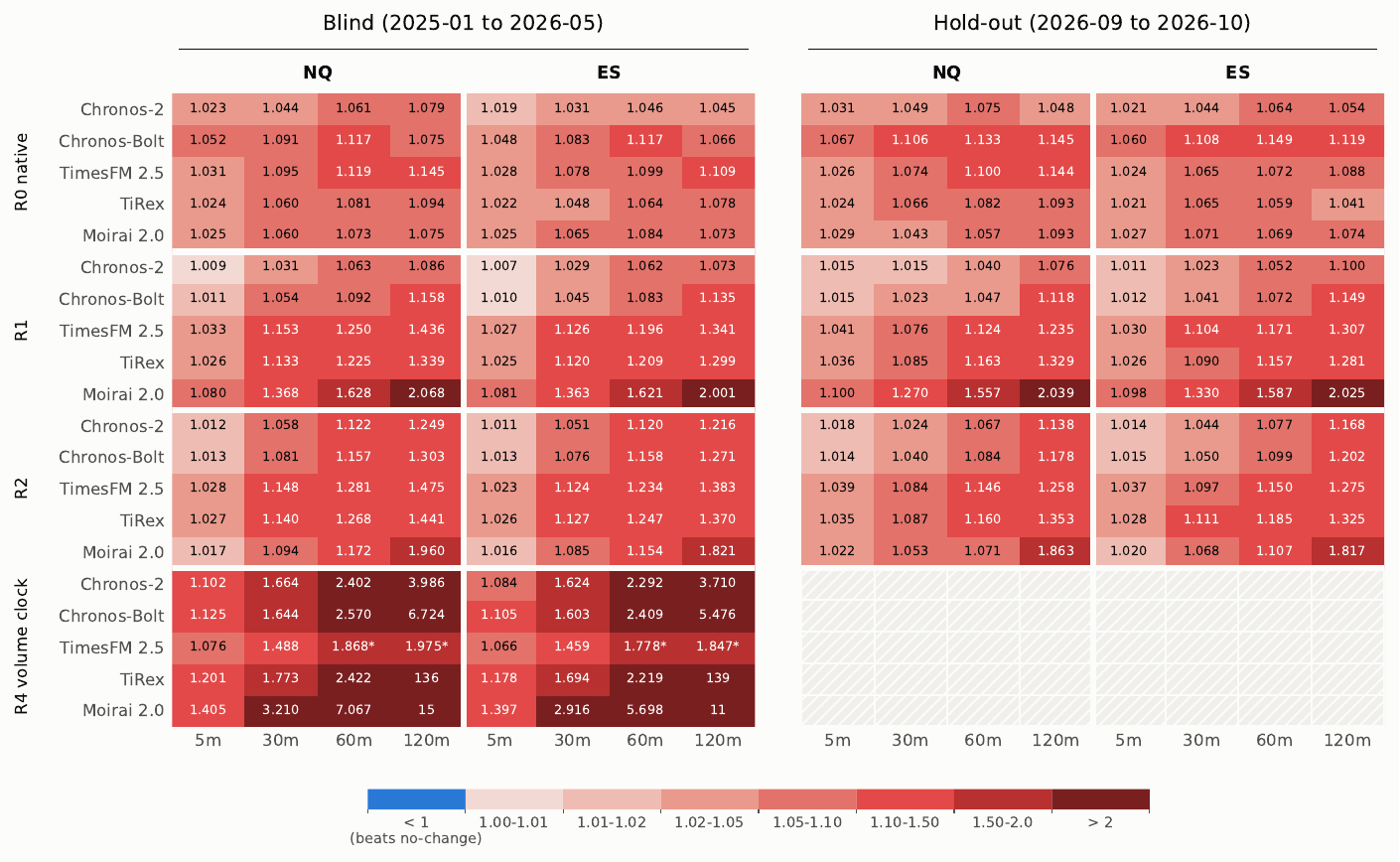}
\caption{MAE ratio vs no-change, model $\times$ input representation $\times$ market $\times$ horizon, for the blind set (left) and the retrospective hold-out (right). Lower is better; below 1 would beat the random walk. No cell is below 1. $^{*}$ partial origin coverage (TimesFM 1,024-step cap on R4); hatched cells were not run (no R4 on the hold-out).}\label{fig:mae}\end{figure}
Figure~\ref{fig:mae} shows every cell and the table summarises them. On Blind, none of the 160 model $\times$ representation $\times$ horizon $\times$ market cells beats no-change, and every one has its own 95\% interval entirely above 1 (separate intervals, not a simultaneous bound). SPA gives p = 1.000 on both markets, in both the available-case and the common-sample analysis, with test statistic T = 0. On the hold-out no point estimate is below 1; the best cell is Chronos-2 R1 ES 5m at 1.0113 [1.0063, 1.0170], with IC -0.016.

{\footnotesize\par\medskip\noindent
\begin{tabularx}{\linewidth}{@{}>{\raggedright\arraybackslash}X>{\raggedright\arraybackslash}X>{\raggedright\arraybackslash}X@{}}
\toprule
 & Blind & Hold-out \\
\midrule
FM cells beating no-change (SMI) & \textbf{0 / 160} & \textbf{0 / 120} \\
SPA, C1 (NQ / ES), both samples & p = 1.000 / 1.000 (T = 0) & not computable (no R4; Appendix~\ref{app:F}) \\
Range of MAE ratios & R0 1.019-1.145; R1/R2 1.007-2.07; R4 1.07-139 & 1.011-2.04 \\
Cells adding over ridge (R1/R2 only) & 0 / 80 & 0 / 80 \\
Lowest per-cell 95\% CI lower bound & 1.0048 (Chronos-2 R1 ES 5m) & 0.973 (14 cells below 1; n $\approx$ 10$\times$ smaller) \\
\bottomrule
\end{tabularx}
\par\medskip}

Validation showed the same pattern. Native-input ratios were 1.02-1.08 for Chronos, TimesFM and Moirai and \textbf{1.72 for Kronos-small} (ES 1.69). Kronos's 20 sampled paths agree in direction far more than outcomes justify (mean $|\text{share up}-0.5|$ $\approx$ 0.39), and its pooled IC is negative on both markets (-0.032 / -0.036). When the forecasts are traded alone, with stops checked on 1-second bars and costs included, no evaluated cell passes the pre-specified economic benchmarks (Appendix~\ref{app:E}).

\subsection{Input representations change magnitude, not skill}\label{sec:5.2}
Returns-based inputs move short-horizon error toward 1 but never across it. R1 or R2 improves on R0 in 13/160 blind cells and 11/120 hold-out cells, mostly at 5 minutes (best: Chronos-2 R1, 1.009 NQ / 1.007 ES). At 30 minutes and longer, R1/R2 are worse than native for 4 of 5 models, for example Moirai R1 at 120 minutes by about 2$\times$. The errors are mostly about forecast size. An exploratory in-validation decoder audit that gives every forecast its in-sample least-squares scale brings the best ratios to 0.9965-1.002 over 120 cells, none reaching the SMI, and leaves almost no directional information. The transformation-aware ridge on the same inputs scores 0.994-1.002, and no foundation-model cell adds over it (0/80 on both samples).

The volume clock is the worst representation as implemented. On Validation 0/40 cells beat no-change or R0, and on Blind the ratios reach 15$\times$ (Moirai) and 139$\times$ (TiRex, 120m). The step mapping is causal but noisy, and the models are rolled out 200-1,100 steps, far beyond their native output lengths. Removing drift leaves the ratios at 1.06-1.6. We report this as a property of the implemented pipeline, not as evidence that event-time information is useless (details in Appendix~\ref{app:E}).

\subsection{Volatility: a registered hypothesis passes, and a simple control explains it}\label{sec:5.3}
On Blind, Moirai 2.0 with R2 input at 5 minutes meets H7's registered rule on both markets:

{\footnotesize\par\medskip\noindent
\begin{tabular*}{\linewidth}{@{\extracolsep{\fill}}lllll@{}}
\toprule
 & QLIKE / HAR & 95\% CI & DM p & 80\% coverage \\
\midrule
NQ & 0.971 & [0.957, 0.986] & 1.2e-4 & 0.815 \\
ES & 0.981 & [0.965, 0.997] & 0.015 & 0.814 \\
\bottomrule
\end{tabular*}
\par\medskip}

Only NQ also meets the stricter volatility SMI; ES at 0.981 misses its 0.98 threshold. The standalone causal EWMA that R2 contains does better than the model (0.955 NQ / 0.968 ES against HAR). Against the EWMA, the model's QLIKE ratio is 1.016 [1.011, 1.022] on NQ and 1.014 [1.007, 1.020] on ES, and both intervals exclude 1. At 30 minutes and longer no model beats HAR (best 1.08-1.12), and on the hold-out no model beats HAR at any horizon. The foundation-model forecast therefore improves on HAR at 5 minutes but is significantly worse than the EWMA control, so there is no evidence of incremental value from the foundation model.

\subsection{Daily direction}\label{sec:5.4}
{\footnotesize\par\medskip\noindent
\begin{tabular*}{\linewidth}{@{\extracolsep{\fill}}lllllllll@{}}
\toprule
 & Market & n & IC & p & Sign-trade & \begin{tabular}[b]{@{}l@{}}Random \\ \relax 95th pct\end{tabular} & Always-long & Verdict \\
\midrule
H9 Blind & NQ & 349 & +0.116 & 0.015 & +\$4,682 & \$16,219 & +\$13,587 & \textbf{fails (trading)} \\
H9 Blind & ES & 332 & +0.034 & 0.27 & +\$351 & \$3,553 & +\$1,212 & - \\
H11 Blind & NQ & 349 & -0.032 & 0.73 & -\$13,576 & \$16,219 & +\$13,587 & \textbf{fails} \\
H9 hold-out & NQ & 119 & -0.008 & 0.53 & +\$1,789 & \$17,168 & -\$2,499 & nothing \\
H9 hold-out & ES & 136 & +0.228 & 0.004 & +\$3,049 & \$1,833 & +\$418 & reported, not claimed \\
\bottomrule
\end{tabular*}
\par\medskip}

Kronos-small's daily rank signal replicated on Blind (IC +0.116, one-sided p = 0.015, against +0.133 on Validation), but trading it earns +\$4,682, below both always-long and the random-direction 95th percentile, so H9 fails its trading condition. On the NQ hold-out the signal vanishes. H11 did not replicate; its sign flipped. The ES hold-out cell clears both bars, but it is a replication-market cell, one of four examined after the primary test had failed twice, so we report it and do not claim it.

\subsection{Controls}\label{sec:5.5}
Placebo windows, built from real 1-minute data with random signs, give the same error and the same forecast magnitude as real data (ratio 0.92-1.14): the models issue directional calls of similar size on noise. Errors on pre-cutoff and post-cutoff years are equal (for example Chronos NQ 30m 1.086 vs 1.084), so there is no sign of memorisation. Shifting Kronos's timestamps by one year or jittering prices by $\pm$1 tick barely changes its forecasts (correlation 0.99-1.00). Kronos trained from scratch on NQ is much less wrong than zero-shot (1.10 vs 1.72; 1.079 after tuning) but still worse than no-change, and Kronos-small fine-tuned on NQ is worse (1.99). None of the pre-listed single changes (volatility scaling, RTH-only context, a 2,048-bar context, agreement filtering of sampled paths) passes.

\subsection{What the study could detect}\label{sec:5.6}
\begin{figure}[t]\centering\includegraphics[width=\linewidth]{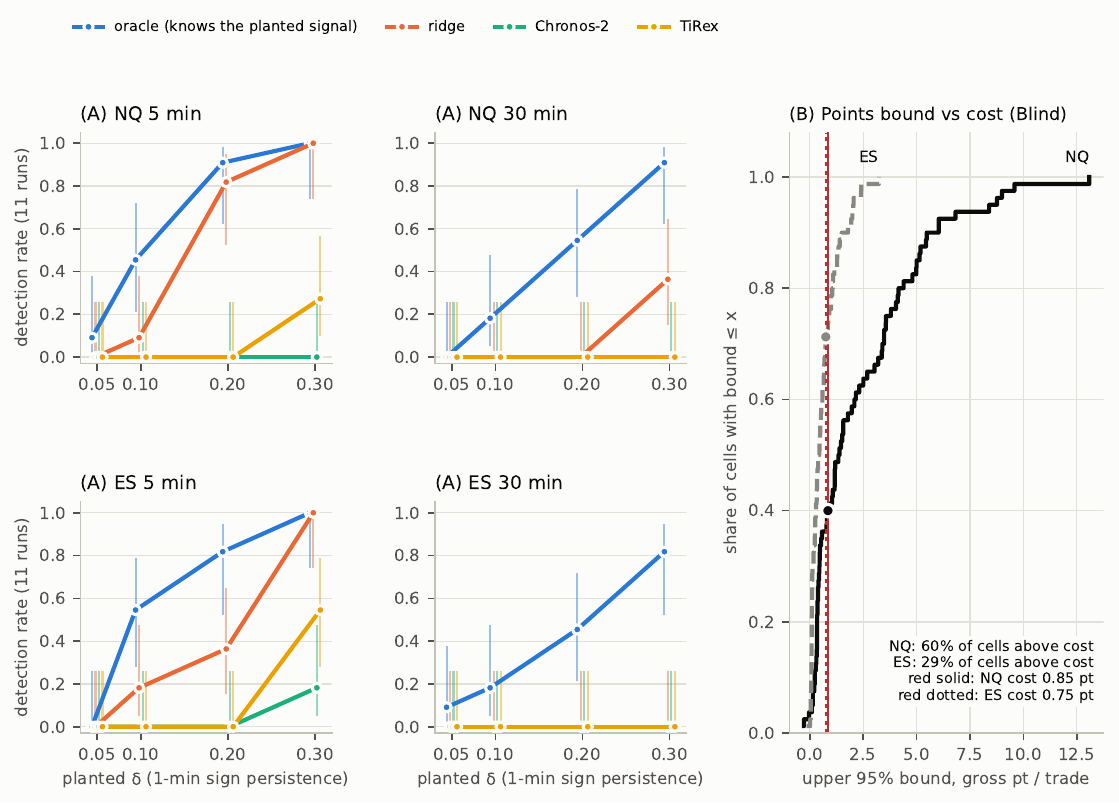}
\caption{What the study could detect. (A) Share of 11 planted-signal realisations in which each forecaster detects the signal (MAE ratio $<1$ AND squared-error DM $p<.05$ AND IC $>0$ with $p<.05$), with 95\% Wilson intervals. (B) Share of blind cells whose per-cell (not simultaneous) upper 95\% bound on gross points per trade lies below a given value, for the policy sign$(f)$ with exit at $t+h$; costs: NQ 0.85 pt, ES 0.75 pt.}\label{fig:power}\end{figure}
The table and Figure~\ref{fig:power} (left) give detection rates over 11 planted-signal realisations of 2,000 origins each, with 95\% Wilson intervals \citep{wilson1927}.

{\footnotesize\par\medskip\noindent
\begin{tabular*}{\linewidth}{@{\extracolsep{\fill}}lllll@{}}
\toprule
 & oracle & ridge & TiRex & Chronos-2 \\
\midrule
NQ 5m, $\delta$ = 0.2 & 10/11 & 9/11 & 0/11 & 0/11 \\
NQ 5m, $\delta$ = 0.3 & 11/11 & 11/11 & 3/11 & 0/11 \\
ES 5m, $\delta$ = 0.3 & 11/11 & 11/11 & 6/11 & 2/11 \\
NQ 30m, $\delta$ = 0.3 & 10/11 & 4/11 & 0/11 & 0/11 \\
\bottomrule
\end{tabular*}
\par\medskip}

Below $\delta$ $\approx$ 0.1 even the oracle rarely detects at this sample size, so the real-data negative is uninformative for edges that small. At $\delta$ = 0.2-0.3, i.e. 70-80\% sign repetition, the tests are capability stress tests rather than realistic alpha; a 20-feature ridge finds the planted structure reliably, while the foundation models mostly miss it. Kronos-small detected nothing on the original grid ($\delta$ $\le$ 0.05, 0/20), which is uninformative because the oracle also fails there.

Because the combined criterion can hide partial recovery, the next table reports how much of a strong signal ($\delta$ = 0.3) the models recover in the saved planted-signal runs (``Detected'' applies the three-condition rule of Section~\ref{sec:4}).

{\footnotesize\par\medskip\noindent
\begin{tabularx}{\linewidth}{@{}>{\raggedright\arraybackslash}Xlll@{}}
\toprule
 & MAE ratio vs no-change & Spearman IC & Detected \\
\midrule
5m, five FMs $\times$ R0/R1 $\times$ NQ/ES (20 cells, one run) & 0.949-0.98 (one cell 1.0003) & 0.19-0.29 & 13 / 20 \\
5m, oracle (NQ / ES) & 0.927 / 0.914 & 0.33 / 0.36 & yes \\
5m, ridge fitted to the process (NQ / ES) & 0.944 / 0.941 & 0.32 / 0.33 & yes \\
Chronos-2 and TiRex, 5m (mean of 11 runs) & 0.969-0.982 & 0.22-0.23 & 11 / 44 runs \\
Chronos-2 and TiRex, 30m (mean of 11 runs) & 1.07-1.12 & 0.03 & 0 / 44 runs \\
Oracle, 30m (mean of 11 runs) & 0.988-0.990 & 0.13-0.15 & 82-91\% \\
\bottomrule
\end{tabularx}
\par\medskip}

At 5 minutes the models recover part of the signal: their MAE improves on no-change by roughly a third to half of the oracle's gain, and their rank association is about two-thirds of the oracle's. Where detection fails at 5 minutes, it is mostly the squared-error significance condition that is not met (for example Chronos-2 R0 NQ: ratio 0.977, DM p 0.30), which we report as a property of the registered combined criterion. At 30 minutes, in the matched pair (the same two models and 11 realisations at both horizons), IC falls to about 0.03 against the oracle's 0.13-0.15, and the forecasts are worse than no-change. Part of this is built into the process. Under the planted Markov sign process, $\mathbb{E}[S_{t+k}\mid S_t]=(2\delta)^k S_t=0.6^k S_t$ at $\delta$ = 0.3, so the predictable part of an h-minute move stays bounded while its noise grows like $\sqrt{h}$. Two caveats apply. Ridge is fitted on series from the same planted process while the foundation models are zero-shot, so the comparison is not a pure capacity comparison. And the evidence is synthetic: it shows neither that the models carry useful ranking information on real returns nor that they generally lose forecasting ability beyond 5 minutes.

\subsection{An economic bound for one fixed policy}\label{sec:5.7}
$\delta$ is not measured in index points, so we also bound the gross points per trade of one policy fixed in advance (Figure~\ref{fig:power}B): trade sign(f) at every blind origin and exit at t+h. The upper 95\% bound (day bootstrap) exceeds the round-trip cost in 60\% of NQ cells and 29\% of ES cells, and the maximum per-cell bound is 13.1 points on NQ against a 0.85-point cost. The bound uses the plan's general round-trip cost for each market (NQ 0.85 pt, ES 0.75 pt); H9 and H11 instead apply their registered rule, written for NQ at 0.85 pt, unchanged to ES. Cells with a lower bound above 0 occur at about the chance rate (3/80 NQ, 4/80 ES). \textbf{The data therefore cannot rule out an economically relevant edge for this policy.} The negative result concerns detected forecasting skill; it is not a proof that no profitable signal exists.

\section{A metric that rewards mean reversion under a random walk}\label{sec:6}
\begin{figure}[t]\centering\includegraphics[width=\linewidth]{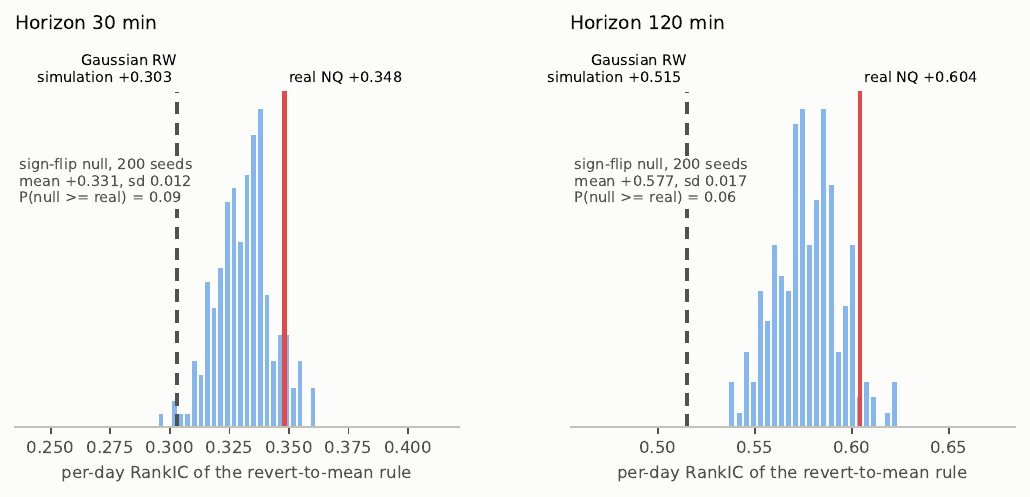}
\caption{Per-day RankIC of the revert-to-mean rule on real NQ (red) against the 200-seed sign-flip null (blue: every 1-minute return given a random sign, volatility kept), and the Monte Carlo mean under a constant-volatility Gaussian random walk (dashed). The null reproduces about 95\% of the observed value and is not rejected at 5\%.}\label{fig:rankic}\end{figure}
On Validation, Kronos-small's per-day RankIC is +0.188 at 30m. Per-day RankIC is here the Spearman correlation between forecasts and outcomes across one day's origins, averaged over days. That looks strong, yet the same forecasts have a pooled RankIC of -0.032 and an MAE of 1.72$\times$ no-change. A trivial rule, f = mean(log P over the last 400 bars) - log P(t), scores \textbf{+0.348} (30m) and \textbf{+0.604} (120m) on this metric, while its pooled RankIC is -0.026 / -0.042. Kronos-small's forecast correlates 0.857 with this rule.

\textbf{Proposition (expected within-day covariance under a random walk).} Let log price be a random walk, $P_j=\sum_{k\le j} e_k$, with independent mean-zero increments of variance $\sigma^2$, and take n origins in one day. Suppose the predictor $x_t$ and the target $y_t=P_{t+h}-P_t$ are linear in the increments, $x=Ae$ and $y=Be$, with $x_t$ using only increments up to t. With $H=I-\mathbf{1}\mathbf{1}^\top/n$,

\[\mathbb{E}[\widehat{\mathrm{cov}}_{\mathrm{day}}(x,y)]=\tfrac{\sigma^2}{n}\operatorname{tr}(A^\top HB),\quad \mathbb{E}[\widehat{\mathrm{var}}_{\mathrm{day}}(x)]=\tfrac{\sigma^2}{n}\operatorname{tr}(A^\top HA),\quad \mathbb{E}[\widehat{\mathrm{var}}_{\mathrm{day}}(y)]=\tfrac{\sigma^2}{n}\operatorname{tr}(B^\top HB).\]

Although $\mathbb{E}[x_t y_t]=0$ at every origin, $\mathbb{E}[\widehat{\mathrm{cov}}_{\mathrm{day}}]$ is in general not zero. It is created entirely by centring within the day, because the day's mean of x contains prices observed after the earlier origins. In particular, for $x_t=-P_t$ with one-step, non-overlapping targets, $\mathbb{E}[\widehat{\mathrm{cov}}_{\mathrm{day}}]=\sigma^2(n-1)/(2n)>0$.

\emph{Proof:} Appendix~\ref{app:A}. The special case was also checked numerically for n = 8, 12 and 24.

The proposition concerns Pearson covariance. The trace identity gives the expected covariance of any linear predictor, but its sign must be evaluated predictor by predictor. The special case proves positivity only for $-P_t$; for the revert-to-400-bar-mean and 60-bar-momentum predictors on our origin grid we evaluated the trace numerically as +9.81$\sigma^2$ / +29.18$\sigma^2$ (revert, 30m / 120m) and -4.38$\sigma^2$ / -18.75$\sigma^2$ (momentum). The proposition is not a closed form for Spearman RankIC, so the rank values below come from Gaussian random-walk Monte Carlo under the same linear model. The mechanism is the small-sample bias long known from the Dickey-Fuller regression with a constant and from Kendall's autocorrelation bias; we claim only its appearance inside a forecast-evaluation metric, not the bias itself.

\textbf{Calibration.} The table compares Gaussian random-walk Monte Carlo (40,000 simulated days, constant volatility, the study's grid) with a volatility-preserving null, real NQ Validation 1-minute returns with random signs (200 seeds, same origins); Figure~\ref{fig:rankic} shows the null distributions.

{\footnotesize\par\medskip\noindent
\begin{tabular*}{\linewidth}{@{\extracolsep{\fill}}llllll@{}}
\toprule
Rule & \begin{tabular}[b]{@{}l@{}}Gaussian RW \\ \relax (no data)\end{tabular} & \begin{tabular}[b]{@{}l@{}}Sign-flip null: \\ \relax mean [min, max]\end{tabular} & Real NQ & Null/real & P(null $\ge$ real) \\
\midrule
revert 30m & +0.303 & +0.331 [+0.295, +0.362] & +0.348 & 95.1\% & 0.09 \\
revert 120m & +0.515 & +0.577 [+0.536, +0.624] & +0.604 & 95.5\% & 0.06 \\
momentum 30m & -0.133 & -0.134 & -0.119 &  &  \\
momentum 120m & -0.326 & -0.282 & -0.280 &  &  \\
\bottomrule
\end{tabular*}
\par\medskip}

With no data at all, the Gaussian simulation gives 85-87\% of the real revert score, and keeping NQ's volatility pattern brings it to about 95\%. The sign-flip test does not reject the directional null at 5\% (p = 0.09 / 0.06). This is not proof that the remainder is zero, so we state the result as reproduction under a null with no directional predictability, not as a causal attribution of every part of the value.

\textbf{Scope and recommendation.} This is \textbf{not} the Kronos paper's headline metric, which correlates predicted and actual price paths across the H steps of one sample. Simulated on random walks (look-back 480, H = 96, 40,000 samples), that metric has expected value zero for causal forecasts (revert +0.004, momentum +0.002, random +0.001), as the sign symmetry of future increments implies. It has a different limitation: a no-change forecast is a constant path, so its correlation is undefined and the metric has no random-walk benchmark. Our level-correlation result makes the same point, since no-change scores 0.99991 (Appendix~\ref{app:B}, row 1). We recommend reporting rank metrics pooled, or across assets at a fixed time, together with a no-change or sign-randomised null, and never reporting within-period rank metrics across a single asset's origins without that null.

\section{Implications for foundation-model evaluation in low-signal domains}\label{sec:7}
Time-series foundation models perform strongly on broad public benchmarks, and nothing here contradicts that. Our evidence concerns one low-predictability domain, intraday index-futures returns, and it gives four concrete reasons not to assume that benchmark-level success carries over. First, the models issue directional calls on noise: on placebo windows they forecast moves of about the same size as on real data (forecast-size ratio 0.92-1.14), with the same error, and Kronos-small's sampled paths agree in direction far more than outcomes justify. Second, they recover strong structure only partly: an 80\%-persistent planted signal is partly recovered at 5 minutes, and recovery deteriorates at 30 minutes under the tested process and model subset. Third, standard metrics look like skill without it: no-change wins on price-level correlation, and within-day RankIC is positive under a random walk. Fourth, simple ingredients explain apparent gains: the volatility gain over HAR is matched by the EWMA the input already contains. Other recent financial evaluations reach compatible conclusions: \citet{konat2026} on fixed income and FX, \citet{noguer2026} on equities, \citet{rahimikia2025} on cross-sectional returns, and \citet{zhong2026} on volatility.

The defensible message is not that these models are without value. It is that adoption in a low-signal domain should follow a domain-specific test against matched nulls, ingredient controls and a magnitude-sensitive loss, rather than benchmark reputation. We do not generalise to other domains, to other financial targets (for example, volatility at longer horizons), or to fine-tuned models.

\section{Limitations and disclosures}\label{sec:8}
The study covers two index futures of one asset class over 2023-2026, and the results may not hold for other assets, frequencies or regimes. Only frozen models were evaluated, except Kronos, which was also trained from scratch and fine-tuned. Part of Blind may predate the training data of some 2025 models; the hold-out does not, but it is short (39 days). The R1/R2 decoder sums per-step medians because the models give per-step quantiles rather than joint paths, so a median-of-sums decoder could not be computed, and the decoder audit covers only linear rescaling. Effects below $\delta$ $\approx$ 0.1 are undetectable even for the oracle at this sample size, and the points bound cannot exclude an edge above cost for the tested policy. Without signed trades, information not contained in prices and volume was not testable.

The pre-registration record is weaker than a third-party registry. An independently implemented code-and-statistics audit, carried out by a separate AI coding agent rather than by human reviewers, rescored every headline table from the saved forecasts, re-ran model inference on about 250 origins per model, and found no critical issue. It raised four disclosure items: earlier plan versions were not kept, so the hash list rests on local session logs; aggregate return statistics for every split were printed before the freeze; Blind was run in several pieces rather than by the single script the plan describes; and the volatility pass was not stated explicitly in the first results log. Appendix~\ref{app:F} gives the audit, every deviation and its timing.

\section{Conclusion}\label{sec:9}
In the locked blind evaluation, five frozen time-series foundation models showed no improvement over a no-change forecast on NQ or ES intraday returns under any tested representation. A retrospective hold-out, scored afterwards as a descriptive check outside the confirmatory family and without R4, agreed. Additional experiments with Kronos (native input, trained from scratch, fine-tuned, and daily) gave no evidence of an economically useful forecasting advantage. The models' apparent successes have simple explanations: price-level correlation, which no-change wins; within-day RankIC, which a random walk inflates for the revert-to-mean predictor on our grid; a volatility gain over HAR, which shows no incremental value over the EWMA its input contains; and a daily rank signal, which does not survive translation into positions or a later sample. Zero-shot models can partially recover deliberately planted short-horizon predictability, yet they show no incremental return-forecasting advantage in the evaluated data, and their apparent success depends on the metric, the horizon and the financial baseline. We therefore recommend that evaluations of forecasting models on financial returns report error against no-change, calibrate rank metrics with a sign-randomised null, include the simple ingredients a representation injects as controls, label the loss behind every test, and state detection limits.

\bibliography{refs}
\bibliographystyle{tmlr}

\clearpage
\appendix
\section{Proof of the proposition (Section~\ref{sec:6})}\label{app:A}
Write $x=Ae$ and $y=Be$ with $e\sim(0,\sigma^2 I)$. The within-day sample covariance with divisor n is $\widehat{\mathrm{cov}}_{\mathrm{day}}=\tfrac1n x^\top Hy=\tfrac1n e^\top A^\top HBe$. For any matrix M, $\mathbb{E}[e^\top Me]=\sigma^2\operatorname{tr}(M)$. This gives $\mathbb{E}[\widehat{\mathrm{cov}}_{\mathrm{day}}]=\tfrac{\sigma^2}{n}\operatorname{tr}(A^\top HB)$, and likewise for the two variances.

For each origin, $x_t$ uses only increments up to t, and $y_t$ only increments in $(t, t+h]$. These are independent with mean zero, so $\mathbb{E}[x_t y_t]=0$, and the uncentred term $\tfrac1n\sum_t \mathbb{E}[x_t y_t]$ vanishes.

The centring term, $-\tfrac{1}{n^2}\sum_s\sum_t \mathbb{E}[x_s y_t]$, is not zero. For a later origin s \textgreater{} t, $x_s$ contains increments in $(t, t+h]$ that also enter $y_t$.

\textbf{Special case.} Let $x_t=-P_t,\ y_t=e_{t+1},\ t=1,\dots,n$. Then $\mathbb{E}[x_s y_t]=-\sigma^2\,\mathbf{1}\{t+1\le s\}$. There are n(n-1)/2 such pairs, so

\[\mathbb{E}[\widehat{\mathrm{cov}}_{\mathrm{day}}] = 0-\tfrac{1}{n^2}\bigl(-\sigma^2\,\tfrac{n(n-1)}{2}\bigr)=\frac{\sigma^2(n-1)}{2n}.\]

Numerical check: research\_notes/rankic\_theory.py, n = 8, 12 and 24.

\section{What each test says (full table)}\label{app:B}
All numbers are from the results log; NQ is primary, ES the replication market.

{\footnotesize
\begin{xltabular}{\linewidth}{@{}l>{\raggedright\arraybackslash\hsize=1.0\hsize}X>{\raggedright\arraybackslash\hsize=0.8\hsize}X>{\raggedright\arraybackslash\hsize=1.6\hsize}X>{\raggedright\arraybackslash\hsize=1.0\hsize}X>{\raggedright\arraybackslash\hsize=0.6\hsize}X@{}}
\toprule
\# & Test (standard metric) & Question it answers & Result in this study & What it can NOT tell us & Source \\
\midrule\endhead
1 & Price-level correlation, $\operatorname{corr}(\hat P(t+h), P(t+h))$ & Does the forecast track the price level? & $\ge$ 0.9996 for every model and horizon, and the no-change forecast scores \textbf{higher} (NQ 30m 0.99991 vs models 0.99969-0.99991) & Anything about skill: the level is dominated by P(t), which every forecast contains & H14 metric bridge (Validation) \\
2 & Within-day RankIC (rank correlation across a day's origins) & Do forecasts rank the day's moves correctly? & Kronos-small +0.188 (30m). A trivial revert-to-mean rule scores +0.348 (30m) / +0.604 (120m). A sign-randomised null reproduces \textasciitilde{}95\% of that (+0.331 / +0.577; not rejected at 5\%, p = 0.09 / 0.06), and a Gaussian random-walk simulation gives +0.303 / +0.515 & Whether a forecaster has directional information: for the revert-to-mean predictor on this within-day origin grid the metric is positive under a random walk (expected covariance in closed form, Section~\ref{sec:6}) & H14; metric artifact \\
3 & Pooled RankIC (all origins together) & Is the forecast's rank related to the outcome's rank? & |RankIC| $\le$ 0.036 for every model intraday; Kronos-small -0.032 & Size of errors, or economic value & H14 \\
4 & MAE ratio vs no-change, day-block bootstrap CI, SMI $\le$ 0.995 with CI upper \textless{} 1 (registered C1/S0) & Does the model forecast the return better than the random walk? & \textbf{Blind 0/160 FM cells beat no-change; hold-out 0/120.} No hold-out point estimate is below 1 (1.011-2.04). Lowest per-cell Blind CI lower bound 1.0048 (not a simultaneous bound) & Whether a smaller-than-detectable signal exists (see rows 9-10) & Blind intraday; hold-out \\
5 & Hansen SPA over the 60-cell primary family (R1/R2/R4 $\times$ 5 FMs $\times$ 4 h) & After multiplicity, is ANY cell better than no-change? & p = 1.000 on NQ and ES, available-case and common-sample & It is ``no superior cell detected'', not proof of no signal & Blind intraday (C1) \\
6 & Input representations R1/R2/R4 vs native R0, and vs ridge on the same inputs & Do input fixes rescue the models? & They shrink forecast size and sometimes reduce error vs R0 (13/160 Blind cells), but 0 beat no-change. Against the transformation-aware ridge (R1/R2 only), 0/80 add over ridge; R4 has no ridge comparator. R4 (volume clock, as implemented) is worst, up to 139$\times$ & Event-time methods in general: R4's mapping back to clock time is implementation-specific & Blind intraday; R4 audit \\
7 & QLIKE vs HAR (H7 rule: QLIKE-DM p \textless{} 0.05, coverage 75-85\%) and vs a causal EWMA control & Do models forecast the SIZE of the move? & \textbf{H7 met on Blind by its registered rule:} Moirai R2 5m, QLIKE/HAR NQ 0.971 [0.957, 0.986], ES 0.981 [0.965, 0.997], coverage 0.815 / 0.814. Only NQ also meets the stricter volatility SMI ($\le$ 0.98). \textbf{Both are significantly worse than the standalone EWMA control} (1.016 [1.011, 1.022] / 1.014 [1.007, 1.020]). Nothing beats HAR on the hold-out & Incremental value of the foundation model: there is no evidence of any over the EWMA its input contains & Blind; hold-out; audit item 4 \\
8 & Daily direction: rank IC + sign-trade vs 200 random-direction draws (registered H9, H11) & Is there a tradeable daily signal? & H9 Kronos-small NQ: IC +0.116, p = 0.015 (replicates Validation), but trade +\$4,682 \textless{} random q95 \$16,219 and \textless{} always-long \$13,587 $\to$ \textbf{fails}. Hold-out NQ IC -0.008. H11 TiRex fade fails (IC -0.032, -\$13,576). (ES hold-out H9 passes both bars: secondary, reported not claimed) & Intraday direction; other daily policies & Blind H9/H11; hold-out \\
9 & Planted-signal power ($\delta$ = 1-minute sign persistence, 11 runs) & Could the pipeline detect a signal if one existed? & $\ge$ 50\% detection: oracle at $\delta$ 0.1-0.2 (5m) / 0.2-0.3 (30m); ridge at 0.2-0.3 (5m), never at 30m. Chronos-2 and TiRex: at most 55\% even at $\delta$ = 0.3 (ES 5m), 0\% at 30m & An effect size in index points: $\delta$ is not a points measure & Power (Validation); Figure~\ref{fig:power} \\
10 & Upper 95\% bound on gross points/trade for the policy sign(f), exit at t+h (descriptive) & Can we rule out an edge above trading cost? & \textbf{No.} 60\% of NQ cells and 29\% of ES cells have a bound above cost (0.85 / 0.75 pt); max per-cell bound 13.1 pt NQ. Cells with lower bound \textgreater{} 0: 3/80 NQ, 4/80 ES ($\approx$ chance) & Any other policy; the bound is per cell, the simultaneous version is wider & Economic bound; Figure~\ref{fig:power} \\
\bottomrule
\end{xltabular}}

\textbf{Reading the table.} The same forecasts look excellent on rows 1-2 and fail rows 4-5. Rows 2 and 7 have simple explanations: a random-walk artifact and an EWMA input. Rows 9-10 set how far the negative result reaches: it excludes the effects the pipeline can detect, not all effects.

\section{Forecast construction and statistical tests}\label{app:C}
This appendix follows the code (tsfm\_infer.py, kronos\_infer.py, score\_rep.py). \textbf{Common to all models.} Origin t is the first print of minute t, and the target is y = log(open(t+h) / open(t)). The context is the last 400 completed 1-minute bars, i.e. rows t-400 .. t-1; context and target windows that cross a roll or a gap longer than 60 minutes are dropped. Every forecast f is a log return anchored at close(t-1) and is scored as a forecast of y, with no re-anchoring.

{\footnotesize\par\medskip\noindent
\begin{tabularx}{\linewidth}{@{}>{\raggedright\arraybackslash}X>{\raggedright\arraybackslash}X>{\raggedright\arraybackslash}X@{}}
\toprule
Model / representation & Model call & Point forecast f \\
\midrule
Chronos-Bolt, Chronos-2, R0 & \texttt{predict\_quantiles(context closes, prediction\_length = h, levels 0.1/0.5/0.9)} & log($q_{50}$ of the step-h close / close(t-1)) \\
TimesFM 2.5, R0 & continuous quantile head, quantile crossing fixed; outputs [mean, q0.1 .. q0.9] & log(q0.5 at step h / close(t-1)) \\
Moirai 2.0, R0 & quantile levels from the module; batch rebuilt per prediction length & log(q0.5 at step h / close(t-1)) \\
TiRex, R0 & 9 quantile levels & log(q0.5 at step h / close(t-1)) \\
Kronos-small (R0 only) & OHLCV candlesticks, per-window normalisation, clip 5; autoregressive sampling T = 1, top-p 0.9, 20 paths & median over paths of log(close at t+h-1 / close(t-1)) \\
Any model, R1 (log returns) & the input is 1-minute log returns $r_{t-400},\dots,r_{t-1}$; h steps of quantiles & f = $\sum_{k=1}^{h} q_{50,k}$ \\
Any model, R2 (EWMA-standardised) & $z_k=r_k/\sigma_k$, with $\sigma_k$ an EWMA ($\alpha$ 0.05) of r$^2$ up to k-1 & f = $\sigma_{t|t-1}$ $\times$ $\sum_k q_{50,k}(z)$, with $\sigma$ held fixed over the horizon \\
Any model, R4 (volume clock) & 400 volume-bar returns closed by t; m = max(1, round(h $\times$ past-60-min bar rate)) steps & f = $\sum_{k=1}^{m} q_{50,k}$; TimesFM drops origins with m \textgreater{} 1,024 \\
\bottomrule
\end{tabularx}
\par\medskip}

\textbf{Comparators.} The no-change forecast is f = 0. The transformation-aware ridge (the R1/R2 comparisons; R4 has no ridge comparator) uses the same transformed series as the model, lags 1-20 plus 5/15/60-step sums, from bars up to t-1 only; its target is y (R1) or y / $\sigma_{t|t-1}$ (R2), and it is fit on 2018-2021 with $\alpha$ chosen on 2022. The validation baselines B2/B3 use returns from P(t) back to closes 1-390 bars earlier, plus bar features. They use P(t) itself and appear only in the validation results of \S\ref{sec:5.1}.

\textbf{Statistical tests.} The day-block bootstrap (block = 1 trading day, 1,000 resamples) gives the intervals for the MAE ratio, IC, ratio vs R0 and ratio vs ridge (paired). Hansen SPA uses the stationary bootstrap over days (mean block 1 day), 1,000 resamples, seed 20261104 and consistent recentring. Diebold-Mariano tests use Newey-West variance with lag 5 and are one-sided; direction tests use squared error and H7 uses QLIKE. Spearman IC p-values are one-sided, and detection rates have Wilson intervals. Each p-value tests a specific loss:

{\footnotesize\par\medskip\noindent
\begin{tabularx}{\linewidth}{@{}>{\raggedright\arraybackslash}X>{\raggedright\arraybackslash}X>{\raggedright\arraybackslash}X@{}}
\toprule
Test & Loss & Used for \\
\midrule
MAE ratio, its bootstrap CI, the SMI & absolute error & every headline direction claim \\
Hansen SPA (C1) & absolute error, $d=|y|-|y-f|$ & the confirmatory test \\
Diebold-Mariano, direction & \textbf{squared error} (Newey-West, lag 5) & descriptive validation p-values (S0) and one condition of the power detection rule \\
Diebold-Mariano, volatility (H7) & \textbf{QLIKE} & H7 \\
\bottomrule
\end{tabularx}
\par\medskip}

The registered tests are kept as they are, and a direction DM p-value is never read as a test of MAE superiority.

\section{Data, samples and protocol chronology}\label{app:D}
\textbf{Daily sample.} At the registered 1-session horizon the daily targets do not overlap, and each observation is a distinct session, so the Spearman p-values use the nominal approximation of independent sessions. Non-overlapping targets do not by themselves guarantee independence, since returns may be serially dependent; H9's failure rests on the trading benchmarks, which do not use this p-value. The daily hold-out is longer than the intraday one because daily TopstepX bars exist from May 2026, while 1-minute history covers only Sep-Oct.

\textbf{Why the daily blind sample has more sessions than the intraday one (NQ 349 vs 333).} The intraday builder drops sessions within one day of a quarterly contract roll (15 sessions), because a context or target window could straddle two contracts; the roll-adjusted daily series keeps them. Two shortened sessions (2025-01-09, 2026-04-03) have daily bars but no regular intraday grid. The last intraday blind day has no daily origin, because its next-session target would cross into the hold-out. ES is the same: 332 vs 316.

\textbf{Which model enters which analysis.}

{\footnotesize\par\medskip\noindent
\begin{tabularx}{\linewidth}{@{}>{\raggedright\arraybackslash}X>{\raggedright\arraybackslash}X>{\raggedright\arraybackslash}X@{}}
\toprule
Analysis & Kronos-small & Five-model panel (Chronos-Bolt, Chronos-2, TimesFM 2.5, Moirai 2.0, TiRex) \\
\midrule
Validation, native input (R0) & yes & yes \\
Representations R1/R2/R4 (validation) & no (tokenizer expects candlesticks) & yes \\
Blind confirmatory family C1 (R1/R2/R4) & no & yes \\
Blind / hold-out per-cell tables (R0-R4 / R0-R2) & no & yes \\
Volatility H7 & no & yes \\
Daily H9 / H11 & H9 & H11 (TiRex) \\
Planted-signal power & original grid only & all five (single run); Chronos-2 and TiRex replicated \\
Training controls (from scratch, fine-tuned) & yes & no \\
\bottomrule
\end{tabularx}
\par\medskip}

\textbf{Protocol chronology} (all times UTC).

{\footnotesize\par\medskip\noindent
\begin{tabularx}{\linewidth}{@{}l>{\raggedright\arraybackslash}X@{}}
\toprule
Stage & What was fixed or seen \\
\midrule
Before 2026-11-02 & Data pipeline and splits defined. The NQ hold-out capture existed (recorded Sep-Oct 2026) but was not used by this study \\
2026-11-02 08:01 & Plan frozen locally (first hash): models, splits, metrics, pass rules. The same day, the data builders printed aggregate statistics (count, mean, sd of returns) for every split, including blind and hold-out (disclosed, Appendix~\ref{app:F}) \\
2026-11-02 - 11-04 & Validation development: baselines, representations, power curves, metric bridge. Amendments logged with hashes (PLAN \S{}13.9-13.11) \\
2026-11-04 03:38-03:49 & Confirmatory family C1, daily hypotheses, missing-data rules and code hashes locked (PLAN \S{}13.12-13.13) \\
2026-11-04 09:02 & Blind opened once and scored with the frozen code: per-origin outcomes first seen here \\
2026-11-04 17:33 & Hold-out opened (amendment PLAN \S{}13.14 logged at the same time) and scored with the same code \\
\bottomrule
\end{tabularx}
\par\medskip}

The hashes are recorded in local session logs, and earlier plan versions were not kept (Appendix~\ref{app:F}).

\section{Additional results}\label{app:E}
\textbf{Trading the forecasts (Validation).} Trading each forecast's sign alone, with stops checked on 1-second bars and costs included, no evaluated foundation-model cell passes the pre-specified economic benchmarks. Most cells lose money. Some are positive but fail the random-direction and always-long baselines; the largest is Chronos-2 NQ 120m, +\$9,099 against a random-direction 95th percentile of +\$13,206 and always-long +\$20,385. Kronos-small is worse than 97-98.5\% of random-direction draws.

\textbf{Decoder audit.} The audit gives every R1/R2 forecast its in-sample least-squares scale. It is an exploratory in-validation linear-rescaling robustness check: least squares is not MAE-optimal, so it is not an upper bound on every linear calibration. The best ratios are 0.9965-1.002 over 120 cells, with 0 reaching the SMI, and de-meaning changes nothing. Under this rescaling the forecasts carry almost no directional information.

\textbf{Volume clock (R4).} Realised future bars over the predicted number of steps m have a median ratio of 0.78-0.93, and only 41-50\% fall within $\pm$25\%: the past-hour rate overstates activity at 60-120 minutes. The models are rolled out 200-1,100 steps, far beyond their native output lengths, and TiRex becomes unstable beyond about 1,000 steps. The failure is not a drift artefact: removing drift leaves the MAE ratios at 1.06-1.6.

\section{Audit, deviations and disclosures}\label{app:F}
\textbf{What the audit was.} It was an independently implemented code-and-statistics audit carried out by a separate AI agent instance (an LLM coding agent). It was given only the code, the plan and the hash record, and it read the results log only after its own checks. It is not human peer review. It covered (i) independent re-scoring of all saved predictions and (ii) a spot check of model inference; it is not a full independent reproduction of every forecast.

\textbf{Re-scoring.} With its own code, the audit rescored every headline table from the saved forecasts: the Blind and hold-out MAE tables to about 1e-14; SPA, with an independent Hansen implementation (T = 0, p = 1); and H9/H11, volatility, power and the points bound. Origin alignment is exact, and a perturbation test found no look-ahead.

\textbf{Re-inference.} CPU re-inference of about 250 origins per model matches the stored forecasts:

{\footnotesize\par\medskip\noindent
\begin{tabularx}{\linewidth}{@{}l>{\raggedright\arraybackslash}Xl@{}}
\toprule
Model & max abs. difference (log-return units) & correlation \\
\midrule
TimesFM & 6e-9 & 1.000000 \\
Moirai & 5e-9 & 1.000000 \\
Chronos-Bolt & 2e-7 & 1.000000 \\
Chronos-2 & 7e-5 (0.7 bp; R4 only, others $\le$ 9e-8) & 0.999988 \\
TiRex & 1.8e-4 (1.8 bp) & 0.99968 \\
\bottomrule
\end{tabularx}
\par\medskip}

For scale, the median |f| of TiRex is about 4.8 bp at native input, and the median |y| on Validation is 4.9 / 12.4 / 17.6 / 24.7 bp on NQ and 3.5 / 9.0 / 12.7 / 17.8 bp on ES at 5 / 30 / 60 / 120 min. The TiRex maximum is therefore not small relative to a single 5-minute forecast, but the forecasts agree almost perfectly overall (correlation $\ge$ 0.9997, sign agreement 98.5-100\%). The non-zero gaps are differences between Apple MPS and CPU arithmetic; repeated CPU runs are identical, and the daily H11 verdict is the same under either set. Kronos was \textbf{not} re-inferred (stopped for compute-sharing reasons), so the Kronos forecasts behind H9 were rescored but not reproduced.

\textbf{Disclosure items.} The audit found no critical issue and four disclosure items.

\begin{enumerate}
\item \textbf{Plan-version record.} The plan file's hash list cannot be checked against stored files, because earlier versions were not kept. Session logs record each hash when it was made, and the amendment fixing the blind-set procedure is logged about five hours before the blind set was opened. These logs are local, not a third-party timestamp.
\item \textbf{Aggregate blind statistics were seen.} The data-building scripts printed the count, date range, mean and standard deviation of returns for every split, including blind and hold-out, on 2026-11-02, before the freeze. No model forecast or per-origin value was seen.
\item \textbf{Blind was run in several pieces,} not by the single script the plan describes. All code that produced the blind forecasts predates the opening, and a post-opening rebuild of the volume bars matched the bars Moirai R4 had used.
\item \textbf{The volatility pass (\S\ref{sec:5.3}) was not stated explicitly} in the first results log. It is now reported as a pass with its explanation.
\end{enumerate}
\textbf{Other deviations,} all logged before the relevant data were seen: the extended power grid (Section~\ref{sec:4}); TimesFM's 1,024-step cap on R4; R4 and SPA not run on the hold-out, because ES lacks 1-second data there and NQ comes from a different feed; and one code change to include the hold-out in H9's bias history. Two further points are noted for clarity: Diebold-Mariano tests use squared errors while the headline ratio uses absolute errors, and the ES cost is 0.85 pt in H9/H11 (the registered H9 rule applied unchanged) but 0.75 pt in the points bound (the plan's general ES cost).

\end{document}